\documentclass{article}

\PassOptionsToPackage{numbers,square,sort&compress}{natbib}
    \usepackage[preprint]{neurips_2022}

\usepackage[utf8]{inputenc} % allow utf-8 input
\usepackage[T1]{fontenc}    % use 8-bit T1 fonts
\usepackage{amsmath}
\usepackage{bm}
\usepackage{wrapfig}
\usepackage{graphicx}
\usepackage{subfig}
\usepackage{caption}
\usepackage{hyperref}       % hyperlinks
\usepackage{cleveref}       % must be loaded after hyperref
\usepackage{url}            % simple URL typesetting
\usepackage{booktabs}       % professional-quality tables
\usepackage{amsfonts}       % blackboard math symbols
\usepackage{nicefrac}       % compact symbols for 1/2, etc.
\usepackage{microtype}      % microtypography
\usepackage[dvipsnames]{xcolor}         % colors
\usepackage[textsize=tiny]{todonotes}
\usepackage{wrapfig}

\crefname{section}{\S}{\S}

\graphicspath{{fig/}}

\title{The Effect of Task Ordering in Continual Learning}

\author{%
  Samuel J.~Bell and Neil D.~Lawrence\\
  Department of Computer Science and Technology\\
  University of Cambridge\\
  Cambridge, UK \\
  \texttt{\{sjb326,ndl21\}@cam.ac.uk} \\
}

\begin{document}

\maketitle

\begin{abstract}
We investigate the effect of task ordering on continual learning performance.
We conduct an extensive series of empirical experiments on synthetic and naturalistic datasets and show that reordering tasks significantly affects the amount of catastrophic forgetting.
Connecting to the field of curriculum learning, we show that the effect of task ordering can be exploited to modify continual learning performance, and present a simple approach for doing so.
Our method computes the distance between all pairs of tasks, where distance is defined as the source task curvature of a gradient step toward the target task.
Using statistically rigorous methods and sound experimental design, we show that task ordering is an important aspect of continual learning that can be modified for improved performance.
\end{abstract}

\section{Introduction}

In continual learning, we seek a model that can learn new things sequentially, without forgetting what has been previously learned.
In practical terms, this is useful for training under constraints on data or compute.
For example, where privacy concerns limit data retention, a continual learning model can be updated using only recent data.
Alternatively, for efficient use of compute resources, we might wish to update a model to reflect a small set of newly-collected data without having to retrain on the entire dataset.

These practical ideals, however, are markedly disconnected from contemporary continual learning benchmarking \citep{Farquhar2018}.
In developing and evaluating novel approaches to continual learning, it is typical to use a dataset such as split or permuted MNIST, CIFAR-10 or CIFAR-100.
In these setups, larger datasets are broken down into smaller tasks to be learned sequentially.
For example, a model may be trained on 0 vs.\ 1 digit classification before progressing to 2 vs.\ 3 and so on.
While previous work has highlighted the crucial role of the inter-task relationship \citep{Ramasesh2020, Bell2021, Nguyen2019, Lee2021}, benchmark results rarely specify the precise order of the task sequence. %, and it is reasonable to assume that where a natural ordering exists (e.g.\ with our MNIST example) it is followed.

In this work we rigorously investigate the effect of task order on continual learning performance.
Conducting an extensive series of empirical experiments involving both synthetic and off-the-shelf data, we ask whether reordering the task set can mitigate catastrophic forgetting.
This question of how ordering affects learning has strong ties with the field of curriculum learning \citep{Bengio2009}, in which individual training examples are ordered by some definition of difficulty.
Training using examples in difficulty order has been shown to result in faster learning and improved generalization error \citep{Hacohen2019, Weinshall2018}.
Here, we consider curricula not of datapoints within a task, but of tasks themselves within a broader sequence.

After finding a significant effect of task order on continual learning performance, we investigate whether task order can be exploited to mitigate catastrophic forgetting.
Taking inspiration from curriculum learning, we look to construct an ordering of tasks---a path through task space---in two ways.
First using information about the task, and second using information about the relationship between the previous task minima and the first gradient step toward the new task.
Finally, we present a simple algorithm for designing such curricula and empirical evidence measuring its efficacy.

Our approach rests on a crucial assumption that is valid in the real-world setting we described above, but rarely accounted for in artificial benchmarks: that we are free to choose the next task.
This is not an unreasonable ask, and forms the basis of active learning (where the most \emph{useful} datapoint is presented next during training), importance sampling, and of course curriculum learning.
Crucially, we always train on the full set of tasks, only choosing between orderings.

Our primary contributions are to demonstrate that task ordering is a significant source of variance in continual learning performance that must be properly accounted for by future work.
Second, we present empirical evidence that this ordering effect can be exploited to significantly increase or reduce catastrophic forgetting.
Third, we introduce a novel method for calculating a path through task space that ties together recent work on loss surface minima curvature \citep{Mirzadeh2020} and on the role of the first gradient step toward the new task \citep{Bell2021}.

\section{Related work}

\subsection{Continual learning}

Forgetting how to perform previously learned tasks has long been an issue for neural networks  \citep{Sutton1986, McCloskey1989}.
Catastrophic forgetting posed a particular challenge for connectionists attempting to model human and animal cognition \citep{Ratcliff1990, French1993}, for there is no natural analogue for such rapid and dramatic performance decline.
In contemporary machine learning, building models that can learn multiple tasks without catastrophic forgetting is termed \emph{continual} or \emph{lifelong} learning.
Recent approaches to enable continual learning on longer task sequences can be broadly categorized into regularization strategies \citep{Kirkpatrick2017, Zenke2017, Kemker2018}; capacity expansion \citep{Goodrich2014}; and replaying past experiences \citep{Robins1995, Kemker2018}.
Empirical experiments have previously revealed that continual learning performance is affected by model architecture and scale \citep{Goodfellow2014}.
Others have shown that semantic task differences, rather than surface transformations, are more likely to cause catastrophic forgetting \citep{Ramasesh2020, Lee2021, Bell2021}.
\citet{Nguyen2019} quantify how the combined complexity of a task set drives forgetting.
Together, these results highlight the importance of the precise inter-task relationship.
Our work expands notions of task similarity to pathways through through sequences.

In \cref{sec:curvature} we make use of \citet{Mirzadeh2020}'s results showing a correlation between the sharpness of a task's minima---bounded by the spectral radius of the Hessian---and propensity to be forgotten.
As the previous task minima is an initialization for the new task, the negative effect of sharpness is  unsurprising when one considers its impact on generalization error more broadly \citep{Hochreiter1997, Li2017, Jastrzkebski2017}.
This aligns with recent work showing that pretrained models find wider minima, and that pretraining partially alleviates catastrophic forgetting \citep{Mehta2021, Ramasesh2022}.
We build upon this idea in section \cref{sec:curvature}, where we explicitly seek out paths of low curvature through task space.
The notion of a continual learning task as a pretraining initialization is particularly helpful as we develop an asymmetric notion of task distance.
After all, a minima for task A may be a good initialization for task B, but the inverse need not be true.

\subsection{Curriculum learning}

Often inspired by the human classroom, curriculum learning refers to training a model with a curriculum of increasing difficulty \citep{Elman1993, Krueger2009, Bengio2009, Kumar2010}.
Typically, the curriculum is over data points, such that individual training examples are sorted into a progression from easy to hard according to some metric.
It is a matter of ongoing debate as to whether curriculum learning improves performance \citep{Bengio2009, Hacohen2019, Weinshall2018, Wu2021}, though it seems increasingly common in practice when training large state-of-the-art models \citep{Wu2021}.
Curriculum learning is closely related to both importance sampling \citep{Schaul2015, Zhao2015, Katharopoulos2018, Baldock2021} and to active learning \citep{Cohn1996, Settles2009}.
In our work, we apply the idea of investigating ordering to whole tasks, rather than individual datapoints.

\subsection{Task ordering}

In their curriculum learning formulation, \citet{Graves2017} consider sub-tasks as ``stepping stones'' towards a final task, and evaluate various techniques to determining their order within a multi-armed bandit setting.
Our work shares this core idea of ordering tasks but applied to the domain of continual learning.
We also differ in our characterization of inter-task distance and our graph-based approach.
Researchers in multi-task learning \citep{Caruana1997} have also investigated curricula over tasks \citep{Pentina2015, Guo2018}.
\citet{Pentina2015} sequentially introduce tasks by order of ``relatedness'', while \citet{Guo2018} gradually blend loss functions across tasks and prioritize exposure to more difficult tasks.
The inspiration for our graph-based approach comes from \citet{Lad2009}, who maximize form autofill accuracy by ordering fields.
\citet{Lad2009} treat tasks as a Tournament, a special case of an directed acyclic graph that expresses approximated preferences over orderings.
Our work is more general than this approach, easily handling undirected and directed graph relations, flexibly allowing for different distance metrics.

\section{Experiments and results}\label{sec:methods}

We present a series of carefully-designed behavioral experiments to test the effect of ordering a sequence of tasks on continual learning performance.
After some preliminary definitions (\cref{sec:definitions,sec:gratings}), we present interwoven methods and results corresponding to the true order in which we conducted this study.
First, we investigate whether, and by how much, task ordering affects continual learning outcomes according to two common metrics (\cref{sec:effect-order}).
Second, we test whether it is possible to exploit this information to choose a good task ordering that improves performance, if we have knowledge about how tasks inter-relate (\cref{sec:navigating}).
Finally, we move beyond this setting and assume no knowledge of tasks, and explore choosing an ordering based loss surface curvatures (\cref{sec:curvature}).
We conclude with further detail on model architectures and training regimes (\cref{sec:further-methods}).

\subsection{Definitions}\label{sec:definitions}

Let $\mathcal{T}_j$ be a task, and $\mathbf{T} = \{\: \mathcal{T}_j \mid 1 \leq j \leq t \:\}$ be an unordered set of $t$ tasks.
Given some ordering over \(\mathbf{T}\), we train a neural network sequentially on each task and measure its performance on the current and prior tasks.
Fixing order for clarity, we denote a model trained on all tasks up to and including $\mathcal{T}_j$ as $\mathcal{M}_{\bm{\theta}_j}$ and its parameters as $\bm{\theta}_j$.
In continual learning, our goal is to learn $\bm{\theta}_{j}$, minimizing the loss function $L_{j}(\bm{\theta}_{j}) \in \mathbb{R}$, without significantly worsening loss on the earlier tasks $L_{i \in \{0,\ldots,j-1\}}(\bm{\theta}_{j})$.
To measure continual learning performance, we follow common practice \citep{Mirzadeh2020, Chaudhry2018a, Chaudhry2018b, Chaudhry2021} and use average accuracy,
\begin{align*}
	\mathbf{A} = \frac{1}{j} \sum^{j}_{i=0} a_i(\theta_{j}) \ ,
\end{align*}
where \(a_i\) is \(\mathcal{T}_i\) validation accuracy, and average forgetting,
\begin{align*}
	\mathbf{F} = \frac{1}{t - 1} \sum^{t-1}_{i=0} \textrm{max}_{j \in \{0, \ldots, t-1\}} (a_i(\theta_{j}) - a_i(\theta_{t})) \ .
\end{align*}

\subsection{Synthetic data with grating classification}\label{sec:gratings}

\begin{wrapfigure}[8]{R}{0.25\textwidth}
    \vspace{-30pt}
    \centering
    \includegraphics[trim={0 0 10cm 0 },clip,width=0.24\textwidth]{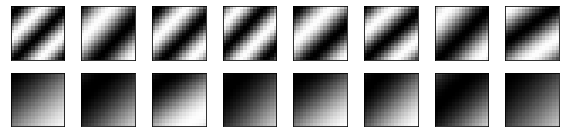}
    \caption{Example binary classification with gratings. Rows are different categories.}\label{fig:task-example}
\end{wrapfigure}

We begin by using synthetic data so that we know the true relationship between tasks.
Our tasks are a binary classification of sinusoidal gratings---as commonly used in psychological experiments---where each grating is defined by its orientation \(\theta\), spatial frequency \(\omega\), and phase \(\phi\).
See \cref{fig:task-example} for an example.
The images have a pixel luminosity at coordinate ($\mathsf{x}$, $\mathsf{y}$) of, following \citep{Goodfellow2012},
\begin{align*}
    l(\mathsf{x}, \mathsf{y} \ ; \ \theta, \omega, \phi) &= \sin(\omega(\mathsf{x}\cos(\theta) + \mathsf{y}\sin(\theta) - \phi)) \, .
\end{align*}
We fully specify each task by its Gaussian data distribution \(X_j\) and decision boundary vector \(\mathbf{b}_j\),
\begin{align*}
    \mathcal{T}_{j} &= (X_j, \mathbf{b}_j) \ , \quad X_j = \mathcal{N}(\bm{\mu}_{j}, \bm{\Sigma}) \ , \\
    \mathbf{x}_{j,i} &= l(\mathsf{x}, \mathsf{y} \ ; \ \theta, \omega, \phi) \ , \quad \left[\theta, \omega\right]^{\top} \sim X_{j} \ , \quad \phi \sim \mathcal{U}(0, 1) \ ,\\
    y_{j,i} &= \textrm{sign}(\left[\theta, \omega\right] \cdot \mathbf{b_j}) \ ,
\end{align*}
where the data distribution is parameterized by its mean vector $\bm{\mu}_j$ and covariance matrix $\bm{\Sigma} = I$.

\subsection{Order significantly affects continual learning performance}\label{sec:effect-order}

As an initial demonstration of reordering, we compare permutations of random triples of tasks with a constant decision boundary,
\begin{align*}
    \mathbf{T}_i &= \{\mathcal{T}_0, \mathcal{T}_1, \mathcal{T}_2\} \ , \quad i < N \ ,\\
    X_j &= \mathcal{N}(\bm{\mu}_{j}, \bm{\Sigma}) \ , \\
    \bm{\mu}_{0} &\sim \mathcal{U}(0, 1) \ , \quad \bm{\mu}_{1} = \bm{\mu}_{0} + \mathbf{o}_1 \ , \quad \bm{\mu}_{2} = \bm{\mu}_{0} + \mathbf{o}_2 \ , \\
    \mathbf{b}_j &= [0, 1]^{\top} \ ,
\end{align*}
where offset \(\mathbf{o}_1\) is a fixed-size positive or negative step in the \(\theta\) dimension, and \(\mathbf{o}_2\) is along \(\omega\).
Thus, given the first task data distribution, the second differs by \(\theta\) only, and the third differs by \(\omega\) only.
We generate \(N=21\) different task sets and compare two different orderings, A and B, of each,
\begin{align*}
    A(\mathbf{T}_i) = \left[\mathcal{T}_0, \mathcal{T}_1, \mathcal{T}_2\right] \quad \textrm{and} \quad B(\mathbf{T}_i) = \left[\mathcal{T}_0, \mathcal{T}_2, \mathcal{T}_1\right] \ .
\end{align*}
We train a model (see \cref{sec:model-arch}) on \(A\) and \(B\) for all \(\mathbf{T}_i\) with a different random seed for each permutation and task set.
We compare \(A\) and \(B\) on average forgetting and average accuracy over the 3 tasks. 

The sample size of \(N=21\) task sets was calculated with a power analysis on pilot data assuming significance threshold \(\alpha=0.05\), power of \(0.8\) and Bonferroni correction for multiple comparisons.
Pilot data was discarded and new task sets generated before full analyses.
This protocol is used consistently in the remainder of our experiments.

Next, we test the extent to which randomly permuting a sequence changes performance, compared with the baseline variation that results from different runs with random seeds.
First, we generalize our random task paradigm to involve sequences of 6 tasks, and to randomly sample both data distribution parameters and decision boundaries,
\begin{align*}
    X_j &= \mathcal{N}(\bm{\mu}_{j}, \bm{\Sigma}) \ , \\
    \bm{\mu}_{j} &\sim \mathcal{U}(0, 1) \ , \\
    \mathbf{b}_{j} &\sim \mathcal{U}(0, 1) \ .
\end{align*}
We generate \(N=14\) task sets and train three orderings: baseline, control and permuted.
Baseline and control are the same random permutations, and permuted is a second random permutation.
We compare the difference in average forgetting and average accuracy between control and baseline (``Control''), and permutation and baseline (``Permutation'').
By doing so, we control for any change in performance induced by training a new model with a random seed, and evaluate the extent to which permutation is a source of variation in relation to this.

\subsubsection{Results}

\begin{figure}[t]
    \centering
    \subfloat[Experiment 1]{
        \includegraphics[trim={0 0 0 0},clip,height=2.5cm]{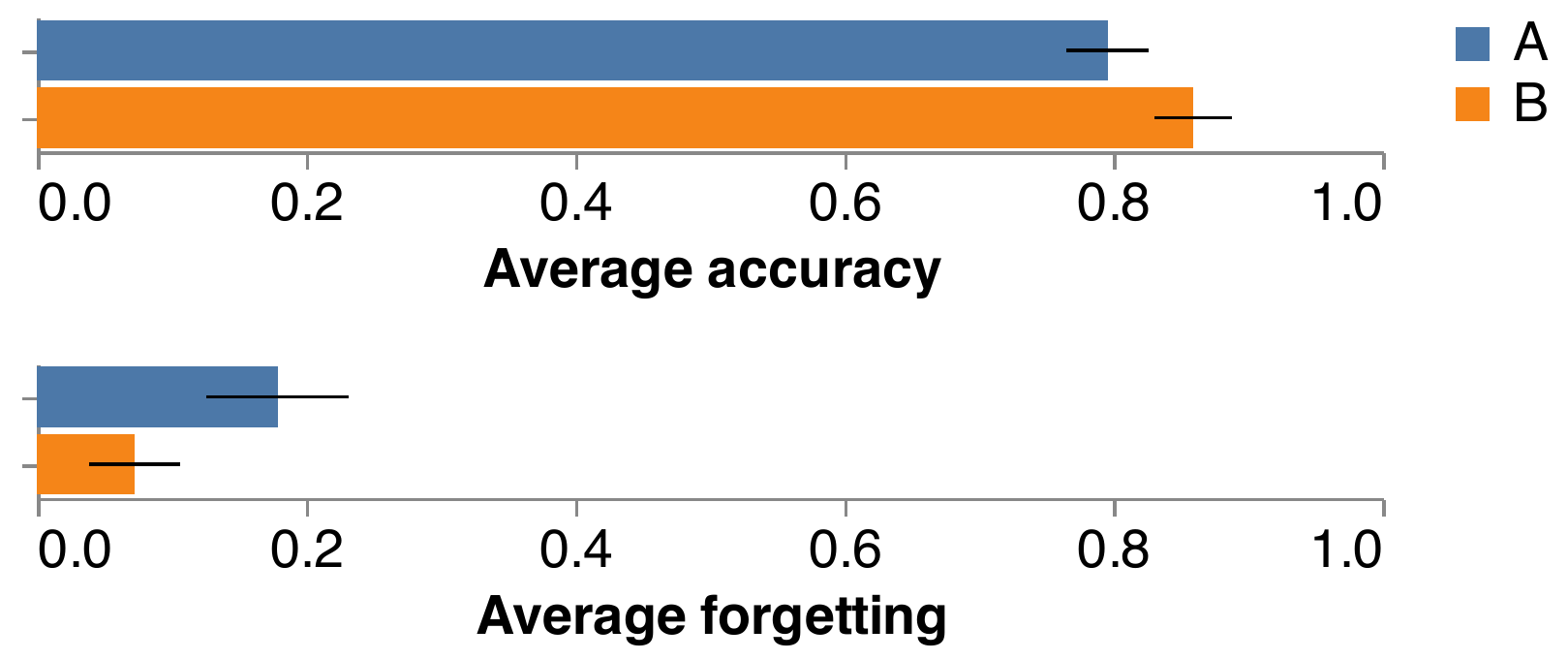}
    }\hfill
    \subfloat[Experiment 2]{
        \includegraphics[trim={0 0 0 0},clip,height=2.5cm]{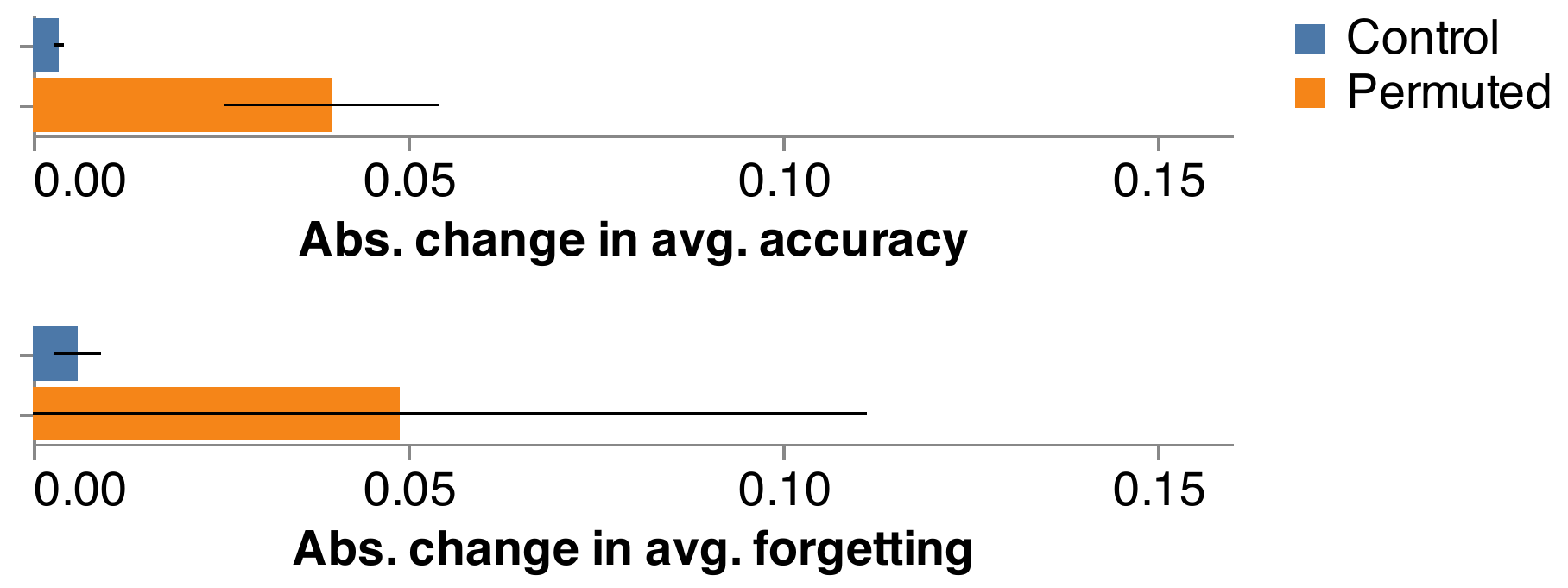}
    }
    \caption{\textbf{(a)} Average accuracy and average forgetting for arbitrary permutations A and B. A trivial reordering results in significantly reduced forgetting and increased accuracy. \textbf{(b)} Abs.\ change in average accuracy and forgetting between two models with different random seeds on the same permutation; and two models with different random seeds and different permutations. Permuting the task set results in significantly more change in both average accuracy and forgetting compared with the random seed control.}\label{fig:experiments-1-and-2}
\end{figure}

Our first results, shown in \cref{fig:experiments-1-and-2}{a}, show that even reordering just two tasks in a triple has a significant effect on average accuracy (\(p < 0.0001\), \(t=7.79\), two-sided independent \(t\)-test) and average forgetting (\(p < 0.0001\), \(t=-7.02\), two-sided independent \(t\)-test).
Order \(A\) has mean 0.11 higher average forgetting than order \(B\), and 0.06 lower average accuracy.

The amount of change induced by permutation is also significant (\cref{fig:experiments-1-and-2}{b}).
In the control condition, retraining a model with a new random seed results in an absolute change in average forgetting of \(0.005\pm0.003\) and in average accuracy of \(0.05\pm0.06\).
There is a significant increase (avg.\ forgetting \(p=0.012\), \(t=2.56\); avg.\ accuracy \(p=0.012\), \(t=2.55\), both one-sided independent \(t\)-tests) between the two conditions, indicating that permutation affects continual learning performance substantially more than variation between model runs.

\begin{figure}[t]
    \subfloat[Baseline - control]{%
        \includegraphics[trim={0 0 0 0},clip,width=1.0\textwidth]{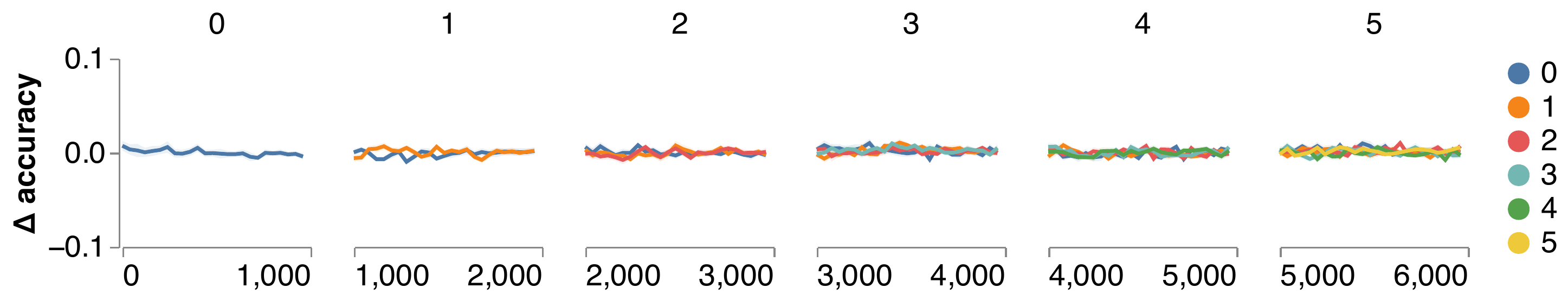}
    }\\
    \subfloat[Baseline - permuted]{%
        \includegraphics[trim={0 0 0 0},clip,width=0.945\textwidth]{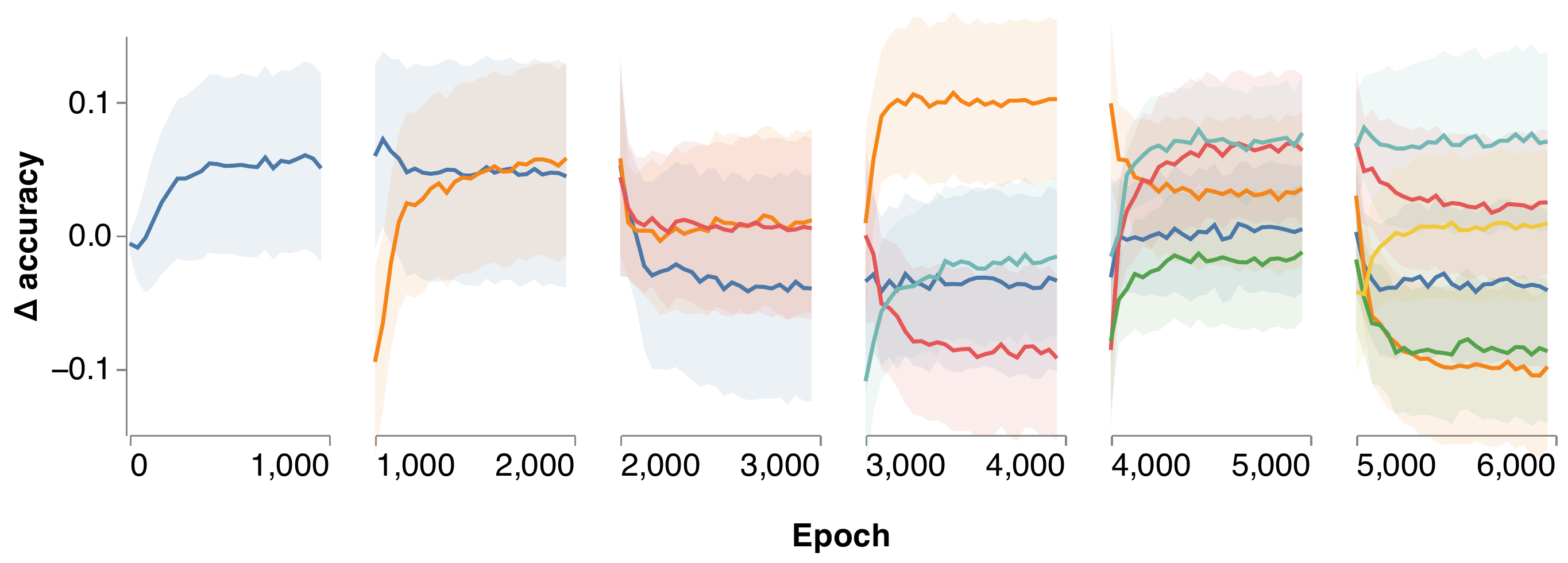}
    }\\
    \caption{Change in validation accuracy as training progresses from task to task for \textbf{(a)} control and \textbf{(b)} permuted conditions. Line color represents the different validation tasks. Shaded area is standard deviation.}\label{fig:accuracy-diff-through-time}
\end{figure}

\Cref{fig:accuracy-diff-through-time} shows the change in accuracy in both conditions as training progresses from task to task.
In \cref{fig:accuracy-diff-through-time}{a} we see that validation accuracy (\emph{not} avg.\ accuracy) does not change between models trained with different random seeds.
\Cref{fig:accuracy-diff-through-time}{b} however shows that training a model in a different order drives an immediate and sustained deviation in validation accuracy from the baseline.

\subsection{Choosing a path through task space}\label{sec:navigating}

\begin{figure}[t]
    \centering
    \subfloat[Distance matrix]{%
        \includegraphics[trim={0 9.4cm 12cm 0},clip,width=0.3\textwidth]{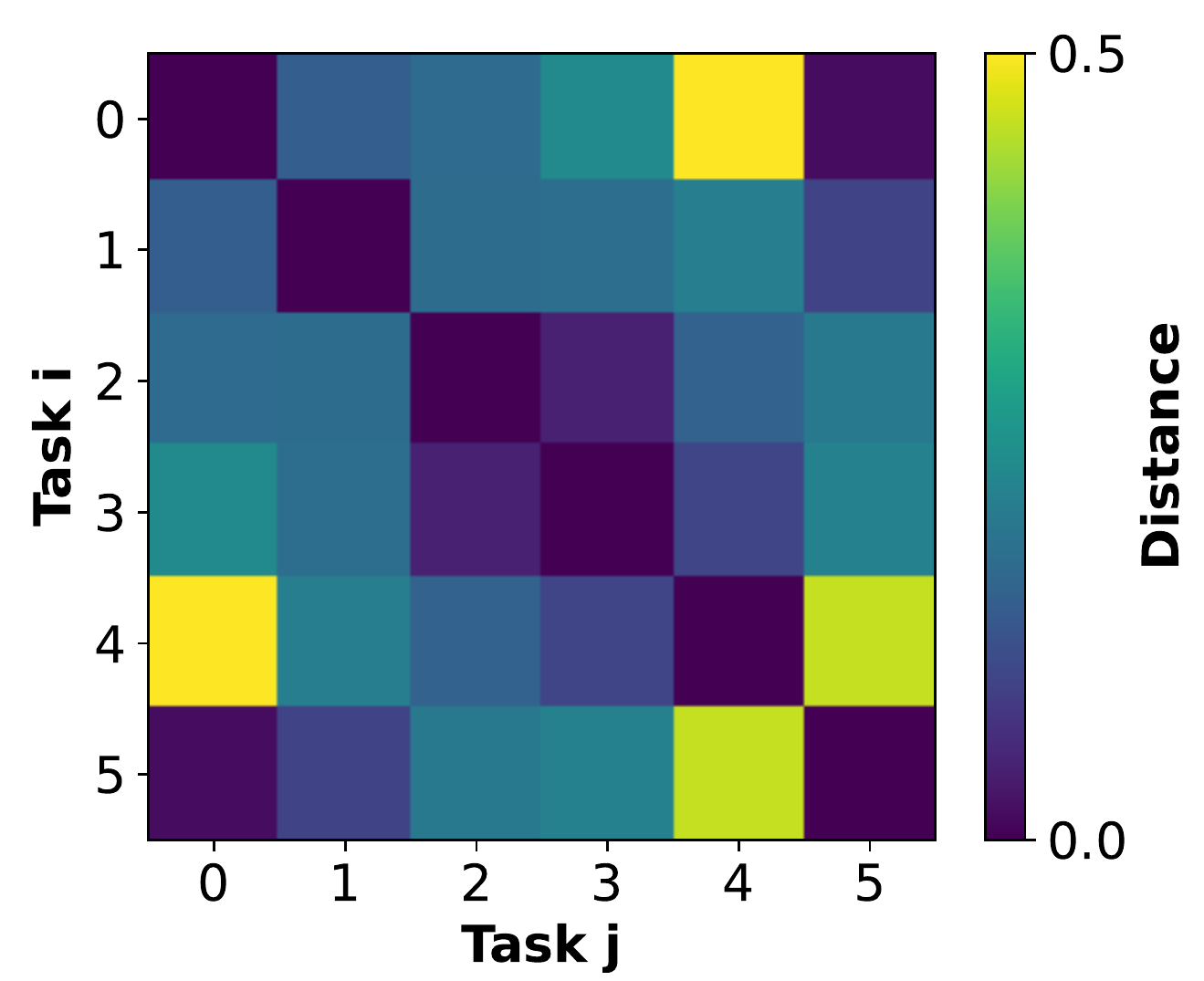}
    }\qquad
    \subfloat[Distance graph]{%
        \includegraphics[trim={14cm 11.2cm 0.5cm 0.5cm},clip,width=0.31\textwidth]{distance-matrix-and-graph-example}
    }\\
    \subfloat[Min sum path]{%
        \includegraphics[trim={1cm 0.6cm 13cm 11.5cm},clip,width=0.32\textwidth]{distance-matrix-and-graph-example}
    }\qquad
    \subfloat[Max sum path]{%
        \includegraphics[trim={13cm 0.6cm 0.5cm 11.5cm},clip,width=0.33\textwidth]{distance-matrix-and-graph-example}
    }
    \caption{\textbf{(a)} Example cosine distances between true task parameters, and \textbf{(b)} their representation as an undirected complete graph. \textbf{(c)} The minimum sum path is the shortest Hamiltonian path through the graph found via exhaustive search. \textbf{(d)} The maximum sum path is the longest route that touches each node exactly once. Edge color corresponds to distance.}\label{fig:example-graph}
\end{figure}

Given knowledge about the tasks in the set, can we devise a route though task space that minimizes forgetting?
Using the synthetic task formulation, recall a task is full specified by \(\bm{\mu}_j\) and \(\mathbf{b}_j\). 
Thus we can calculate task similarity, and choosing cosine distance we produce a distance matrix (\cref{fig:example-graph}{a}) and corresponding weighted graph (\cref{fig:example-graph}{b}).
We hypothesize that taking the shortest path through task space will minimize average forgetting.
We test this by computing three orderings for each task set: a random baseline, a minimum sum path and a maximum sum path (see \cref{fig:example-graph}{c--d}).
We include the counter-intuitive maximum sum path motivated by research in curriculum learning showing that an \emph{anti}-curriculum---hardest to easiest---can be effective.
For the small task sets in our work it is possible to compute these paths exhaustively, though future work may need to adopt more efficient approaches to the traveling salesman problem.
We train a new model on each permutation of \(N=20\) task sets.

\subsubsection{Results}

In \cref{fig:all-tasks-ordered}{a} we compare a minimum sum path (i.e.\, the shortest Hamiltonian path), the maximum sum path, and a randomly chosen path by average forgetting.
We find a significant decrease (\(p=0.008\), \(t=-2.66\), one-sided paired \(t\)-test) in average forgetting when taking the max sum path compared to the min sum path, corresponding to a mean decrease of \(-0.056\).
We find no significant difference between the random path and either the max sum (\(p=0.09\), \(t=-1.39\), one-sided paired \(t\)-test) or min sum paths (\(p=0.05\), \(t=1.74\), one-sided paired \(t\)-test).
In short, choosing the max sum path decreases catastrophic forgetting, and choosing the min sum path increases it, though in neither case enough to be significantly different than a random path.
We note that a random path is quite a stringent test to pass, as the random path condition could easily include paths of similar length (or shortness) as the max sum (or min sum) conditions, and thus we expect the resulting distributions of average forgetting to have substantial overlap.

\subsection{Choosing a path from curvature}\label{sec:curvature}

In a realistic scenario, one is unlikely to have access to the true task parameters that are required to calculate a path through task space.
Here we evaluate an alternative approach that makes use of information about the loss surface around task minima and the direction of travel taken by SGD between tasks.
Previous work has shown that shallower minima are linked to both lower generalization error \citep{Hochreiter1997, Li2017, Jastrzkebski2017} and lower catastrophic forgetting \citep{Mirzadeh2020}.
Others \citep{Bell2021} have suggested that while the former task minima is important, the direction of travel when optimizing for the new task should also be taken into account.
Here, we elegantly combine these two ideas and consider the curvature of the former task, i.e.\ the Hessian of the former loss, \emph{in the direction} of the new task, i.e.\ the gradient of the new loss.
To do this, we first train a model from a shared initialization, on each task in the task set individually.
Thus for each task \(\mathcal{T}_j\) we have a model $\mathcal{M}_{\bm{\theta}_j}$ trained on \(\mathcal{T}_j\) \emph{only}.
After calculating the Hessian around this minima 
\begin{align*}
    \mathbf{H}_j = \nabla^{2}_{\bm{\theta}_{j}} L_{j}(\bm{\theta}_{j}) \ ,
\end{align*}
we evaluate each model on a batch from every other task \(\mathcal{T}_k, k \neq j\) and keep track of the gradient of the loss w.r.t.\ the parameters \(\bm{\theta}_j\),
\begin{align*}
    \mathbf{g}_k = \nabla_{\bm{\theta}_{j}} L_{k}(\bm{\theta}_{j}) \ . \\ 
\end{align*}
Finally, our measure of curvature \(c(j, k)\) in the direction of the new task \(\mathcal{T}_k\), starting from a model trained on \(\mathcal{T}_j\) as,
\begin{align*}
    c(j,k) = \mathbf{g}_k^{\top} \mathbf{H}_j \mathbf{g}_k \ ,
\end{align*}
which can be efficiently calculated even for large models by using Hessian-vector product \citep{Christianson1992} to avoid computing the full Hessian matrix,
\begin{align*}
    \mathbf{H}_j \mathbf{g}_k = \nabla_{\bm{\theta}_{j}} \left[\mathbf{g}_j^{\top} \mathbf{g}_k\right] \ .
\end{align*}
We note that \(c\) is not symmetric, and moving from one task to another may be low curvature but the inverse high.
Thus we have a directed distance graph, rather than the undirected graph in \cref{fig:example-graph}{b}, but otherwise our approach is the same.
We compute a baseline, minimum sum and maximum sum path through curvature and compare models trained on each ordering by average forgetting.
We use a sample of \(N=17\) task sets.

We also test whether our results with min and max curvature paths generalize to more realistic settings by testing on split MNIST \citep{LeCun2010, Zenke2017}.
We draw \(N=10\) sets of 5 binary classification tasks, where each task is a random pair of digits 0--9.
Images are resized to $16\times16$ pixels using bilinear interpolation to match the models we use in the previous experiments.

We then test the generalizability of our results to different models by testing ResNet-18 \citep{He2016} on a version of split CIFAR-10 \citep{Krizhevsky2009}.
Following the same principle as above, we draw \(N=10\) sets of 5 binary classification tasks, where the two classes are sampled randomly.
We use the original $32\times32$ RGB images.
We also test ResNet on longer sequences drawn from CIFAR-100 \citep{Krizhevsky2009}, drawing \(N=18\) sets of 10 binary tasks with randomly drawn pairs of classes.

\subsubsection{Results}

\begin{figure}[t]
    \centering
    \subfloat[Synth.\ true]{\
        \parbox{0.13\textwidth}{
            \centering
            \includegraphics[trim={0 0 4.5cm 0},clip,height=5cm]{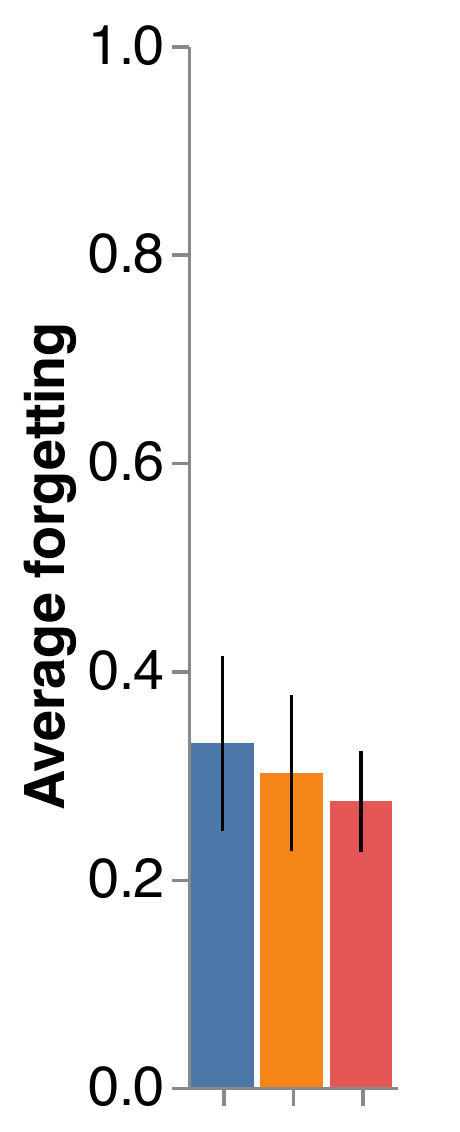}
        }
    }\hfill
    \subfloat[Synth.\ curvature]{\
        \parbox{0.165\textwidth}{
            \centering
            \includegraphics[trim={0 0 4.5cm 0},clip,height=5cm]{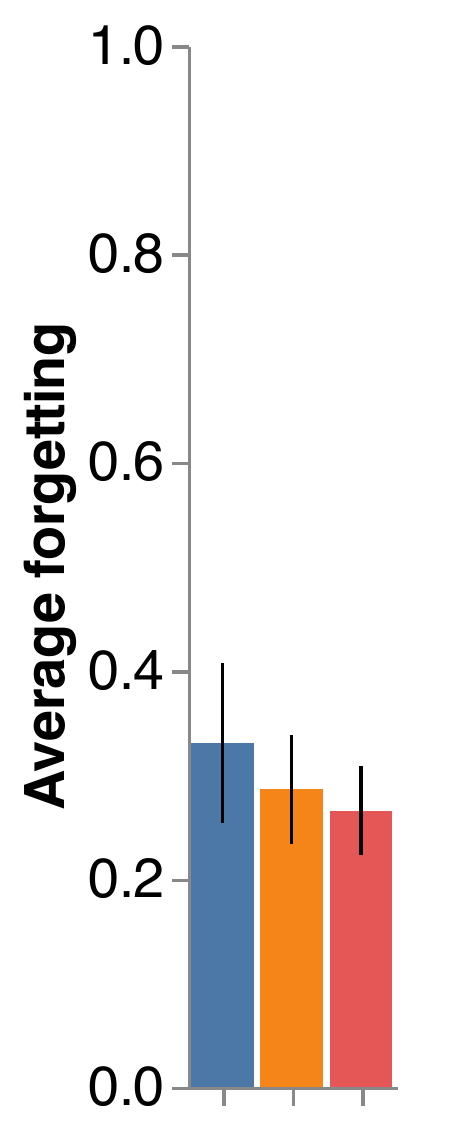}
        }
    }\hfill
    \subfloat[MNIST curvature]{\
        \parbox{0.18\textwidth}{
            \centering
            \includegraphics[trim={0 0 4.5cm 0},clip,height=5cm]{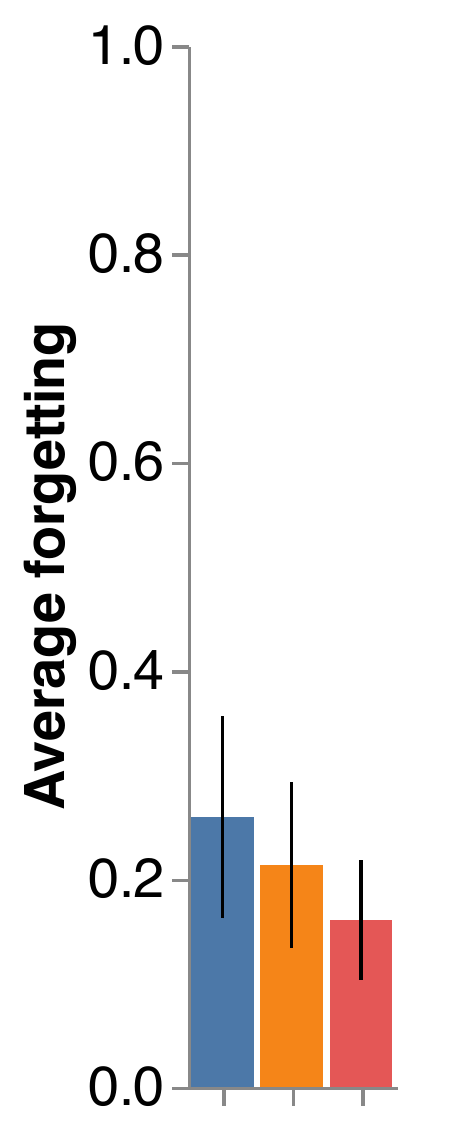}
        }
    }\hfill
    \subfloat[CIFAR10 curvature]{\
        \parbox{0.20\textwidth}{
            \centering
            \includegraphics[trim={0 0 4.5cm 0},clip,height=5cm]{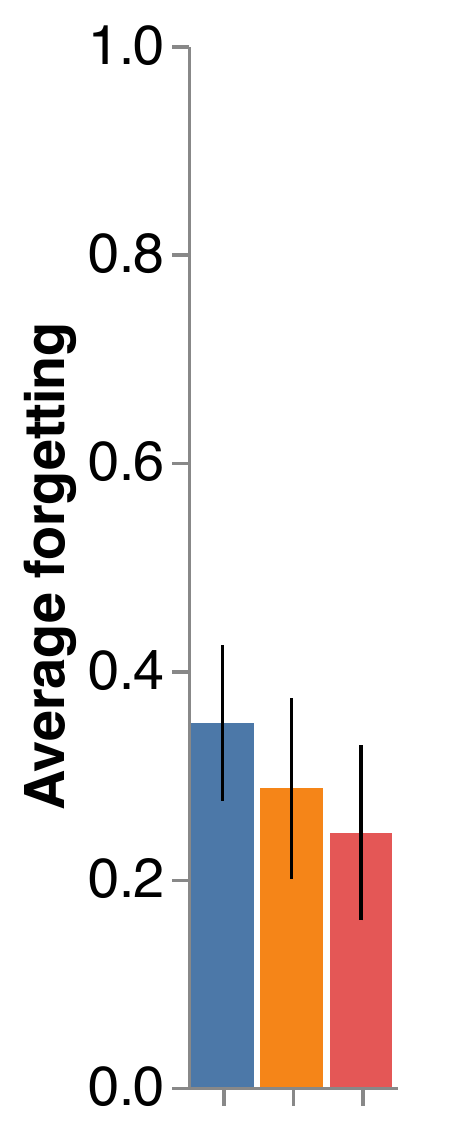}
        }
    }\hfill
    \subfloat[CIFAR100 curvature]{\
        \parbox{0.21\textwidth}{
            \centering
            \includegraphics[trim={0 0 0 0},clip,height=5cm]{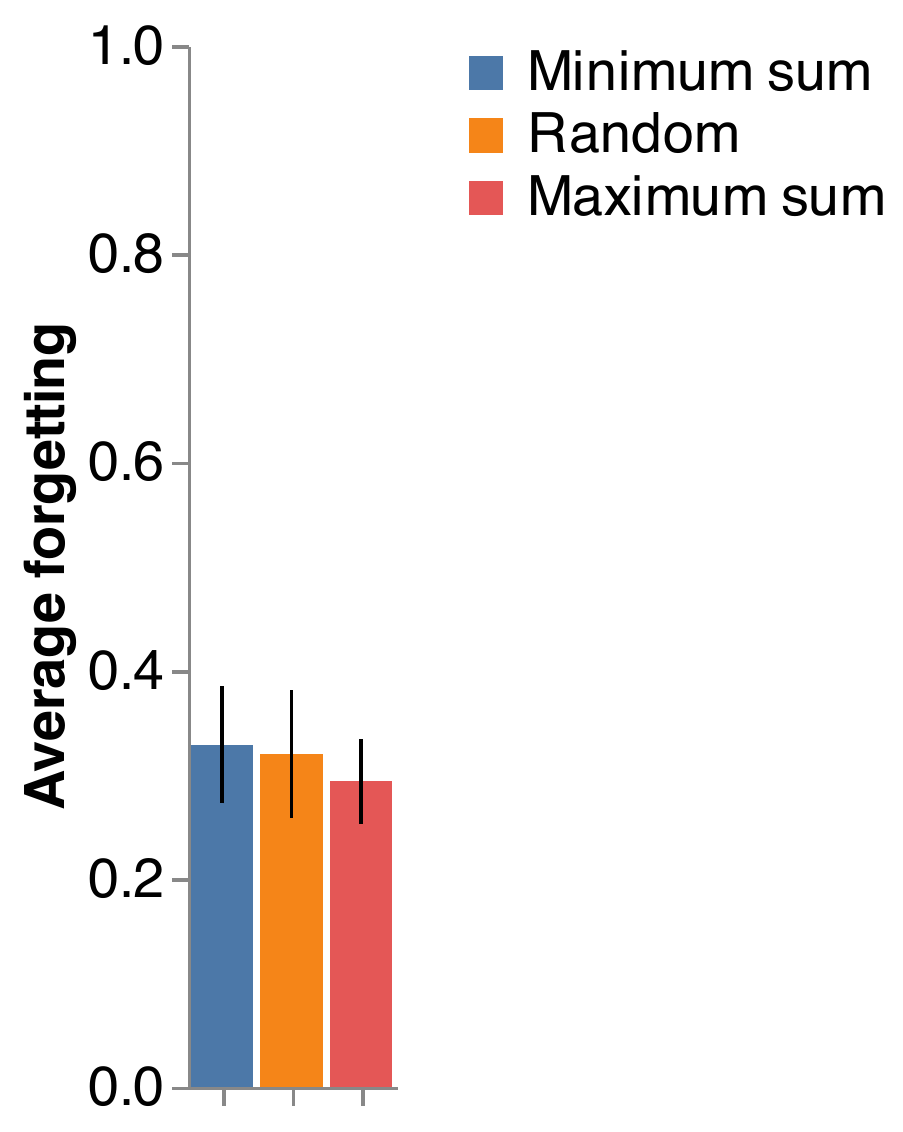}
        }
    }
    \caption{Average forgetting for minimum sum, random and maximum sum paths. Max sum path consistently associated with lowest forgetting.}\label{fig:all-tasks-ordered}
\end{figure}

%Experiments 5-8 tackle the more realistic scenario where we lack information about the inter-task relationship.
Across synthetic gratings, MNIST, CIFAR-10 and CIFAR-100 we show that our curvature-based paths have a significant effect on catastrophic forgetting.
For a two-layer ReLU net on the gratings task (\cref{fig:all-tasks-ordered}{b}) we report a significant decrease (\(p=0.005\), \(t=-2.89\), one-sided paired \(t\)-test) in average forgetting for the max sum path compared to the min sum path, with mean decrease \(-0.065\).
We find no significant differences after correction for multiple comparisons between the min and random (\(p=0.02\), \(t=2.16\), one-sided paired \(t\)-test) nor max and random paths (\(p=0.04\), \(t=-1.81\), one-sided paired \(t\)-test).
For the same model on split MNIST (\cref{fig:all-tasks-ordered}{c}) we find a significant decrease in average forgetting for the maximum sum path versus both the minimum sum path (\(p=0.0012\), \(t=-4.19\), one-sided paired \(t\)-test) and the random path (\(p=0.0096\), \(t=-2.84\), one-sided paired \(t\)-test).
This is a mean decrease in average forgetting of -0.052 versus a random baseline.

Turning to ResNet, on CIFAR-10 (\cref{fig:all-tasks-ordered}{d}) we again find a significant decrease in average forgetting between the maximum and minimum sum paths (\(p=0.0005\), \(t=-4.71\), one-sided paired \(t\)-test), corresponding to a sizeable mean decrease in average forgetting of -0.011.
However, we find no significant difference (after correction) between the random path and either the min (\(p=0.03\), \(t=2.16\), one-sided paired \(t\)-test) or max (\(p=0.05\), \(t=-1.78\), one-sided paired \(t\)-test) sum paths.
These results are consistent with ResNet on CIFAR-100 (\cref{fig:all-tasks-ordered}{e}), again with a significant decrease between max and min sum paths (\(p=0.0077\), \(t=-2.69\), one-sided paired \(t\)-test) and mean decrease in average forgetting of -0.035.
We find no significant difference between either the max (\(p=0.024\), \(t=-2.13\), one-sided paired \(t\)-test) and min (\(p=0.28\), \(t=0.61\), one-sided paired \(t\)-test) sum paths and the random path.

To summarize, we find a consistent and significant difference across all experiments between the maximum and minimum sum paths, i.e.\ taking the path of maximum curvature through task space results in the lowest average forgetting.
However, this is only sufficient to outperform the random baseline in one of four cases, with the ReLU net on split MNIST.
See \cref{table:curvature-results} for a summary.

\begin{table}[t]
    \centering
    \begin{tabular}{llrrr}
    \toprule
    \multicolumn{2}{l}{} & \multicolumn{3}{c}{\textbf{Average forgetting}}  \\
    \cmidrule(l){3-5}
    Model       &       Dataset     &   Max\(-\)min sum   &   Random\(-\)min sum              &   Random\(-\)max sum \\
    \midrule
    Relu Net    &       Gratings    &   \textbf{*-0.065}  & 0.047                             &  \textbf{-0.021}     \\
    Relu Net    &       MNIST       &   \textbf{*-0.098}  & 0.046                             &  \textbf{*-0.053}    \\
    ResNet      &       CIFAR-10    &   \textbf{*-0.105}  & 0.063                             &  \textbf{-0.043}     \\
    ResNet      &       CIFAR-100   &   \textbf{*-0.035}  & 0.009                             &  \textbf{-0.026}     \\
    \bottomrule
    \end{tabular}
    \caption{Mean change in average forgetting for max sum vs.\ random vs.\ min sum paths. \textbf{Entries in bold} represent decreased forgetting i.e.\ improved performance; asterisked are significant differences.}\label{table:curvature-results}
\end{table}

\subsection{Additional methods}\label{sec:further-methods}

\subsubsection{Model architectures}\label{sec:model-arch}

In all experiments with synthetic data and split MNIST we use two-layer ReLU networks.
Input images are flattened and batch normalized before the forward pass.
Hidden layer width is 256, and for binary classification we use a sigmoid-transformed scalar output.
Model parameters are randomly initialized from $\mathcal{U}(-16, 16)$.

Experiments with ResNet-18 use the TorchVision \citep{Paszke2017} implementation.
Model parameters are initialized according to \citet{He2016}.

\subsubsection{Model training}\label{sec:model-training}

At the start of each run on an ordered task set, parameters are randomly initialized.
For subsequent tasks model parameters for task $j$ are set to $\bm{\theta}_{j-1}$.
The network is then trained sequentially on each task according to the ordering, using batches of 128 images, for 1,000 epochs with the ReLU net, and 3,000 epochs for ResNet.
Networks are trained to minimize the cross-entropy loss using SGD with learning rate 0.001.
No other model or optimization hyperparameters were tuned.

\section{Discussion}\label{sec:discussion}

Our results have demonstrated that the order in which tasks are presented significantly affects continual learning performance.
We have also shown that it is possible to choose a route through task space that mitigates catastrophic forgetting, and presented a simple approach for doing so.
In the absence of ground-truth information about each task, one can efficiently compute the curvature in the direction of the upcoming candidate task, and choose a path that maximizes this curvature overall.

Taking the route of maximum curvature is the opposite result to our original hypothesis, though this approach consistently outperforms minimum curvature and occasionally outperforms the random baseline.
We consider the max sum path as analogous to the ``anti-curriculum'' approach explored in curriculum learning \citep{Bengio2009, Weinshall2018, Hacohen2019, Wu2021} where training progresses from difficult to easier examples.
Our intuition for the utility of the maximum sum path is also informed by our experiment with true task distance, where taking the longest path between tasks minimizes forgetting.
In both cases, we posit that taking the longest path maximizes separation between each task, potentially minimizing representational overlap, which has previously been linked to catastrophic forgetting \citep{French1993, Ramasesh2020, Ramasesh2022}.
An exciting area for future research is to characterize how representations change under the conditions of different orderings.

\begin{wrapfigure}[20]{R}{0.3\textwidth}
    \vspace{-10pt}
    \centering
    \includegraphics[trim={0 0 0 0},clip,height=5cm]{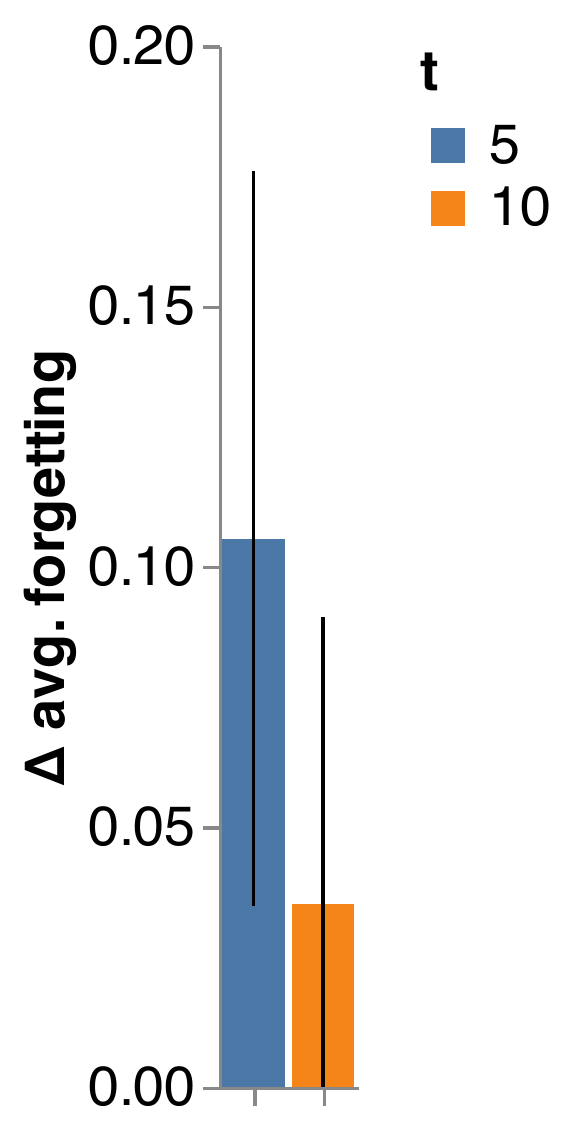}
    \caption{Change in average forgetting between min and max sum paths, for short and long sequence lengths. Error bars are standard deviation.}\label{fig:sequence-length}
\end{wrapfigure}

We believe we are the first to investigate the effect of task ordering in the continual learning setting.
However, this initial investigation comes with a number of limitations that should be considered when generalizing our claims to other settings.
We have chosen the more challenging \citep{Farquhar2018} single-headed model though impact on multi-headed models remains open for investigation, as is a move beyond supervised learning, and specifically binary classification that is covered here.
One significant limitation that we hope to address is that contrary to our hypothesis, increasing the size of the task set decreases the effect of ordering, according to a comparison between ResNet on CIFAR-10 and CIFAR-100 with \(t=5\) and \(t=10\) respectively (see \cref{fig:sequence-length}).
These limitations notwithstanding, our work demonstrates that ordering is an important factor which future research can both capitalize upon and extend to longer sequences.

We have shown that it is possible to exploit information about the task (e.g.\ true parameters) and about the model (e.g.\ minima Hessian, gradient step) to choose an effective route through task space.
Future work may explore alternative methods for characterizing inter-task distance.
For example, a number of measures of example difficulty have been proposed in the curriculum learning and importance sampling literature.
\citet{Bengio2009} treat proximity to the decision margin as a marker of difficulty, or in a synthetic setup the number of ``noisy'' inputs, while others rank examples by gradient variance \citep{Agarwal2020, Katharopoulos2018}.
\citet{Weinshall2018} also consider distance to margin, though using the final layer representation of a pretrained network as input.
Extending this, a measure of representational similarity over the entire dataset, for example Centered Kernel Alignment (CKA) \citep{Kornblith2019}, may be an appropriate method for computing inter-task distance.

\citet{Tsvetkov2016} use outer-loop optimization to develop curricula for learning better performing word embeddings, and \citet{Graves2017} choose between sub-tasks with an multi-armed bandit.
Both ideas could be readily applied to choosing a task order for continual learning.
Our work approaches navigation via a graph structure, which limits our analysis to consideration of discrete tasks.
In a deployed continual learning scenario the task boundary is likely to be porous and even unlabeled \citep{Farquhar2018}.
As such, we also foresee a connection between task ordering and recent research on meta-learning for continual learning \citep{Javed2019, Beaulieu2020}.
Moving beyond discrete tasks blurs the distinction between inter-task curricula, as studied here, and the conventional curricula of training samples, opening the door to new avenues for future research.

%\item Effect of dataset scale, task complexity, network capacity, pretraining

% \begin{ack}
% Use unnumbered first level headings for the acknowledgments. All acknowledgments
% More information about this disclosure can be found at: \url{https://neurips.cc/Conferences/2022/PaperInformation/FundingDisclosure}.
% Do {\bf not} include this section in the anonymized submission, only in the final paper. You can use the \texttt{ack} environment provided in the style file to autmoatically hide this section in the anonymized submission.
% Acknowledgements here.
% \end{ack}

%%%%%%%%%%%%%%%%%%%%%%%%%%%%%%%%%%%%%%%%%%%%%%%%%%%%%%%%%%%%

% \nocite{*} 

\small
\bibliography{curricula-main}

\begin{thebibliography}{53}
\expandafter\ifx\csname natexlab\endcsname\relax\def\natexlab#1{#1}\fi
\expandafter\ifx\csname url\endcsname\relax
  \def\url#1{{\tt #1}}\fi

\bibitem[Farquhar and Gal(2019)]{Farquhar2018}
Sebastian Farquhar and Yarin Gal.
\newblock Towards robust evaluations of continual learning.
\newblock 2019, arXiv:1805.09733.

\bibitem[Ramasesh et~al.(2020)Ramasesh, Dyer, and Raghu]{Ramasesh2020}
Vinay~V. Ramasesh, Ethan Dyer, and Maithra Raghu.
\newblock Anatomy of catastrophic forgetting: Hidden representations and task
  semantics.
\newblock 2020, arXiv:2007.07400.

\bibitem[Bell and Lawrence(2021)]{Bell2021}
Samuel~J. Bell and Neil~D. Lawrence.
\newblock Behavioral experiments for understanding catastrophic forgetting.
\newblock 2021, arXiv:2110.10570.

\bibitem[Nguyen et~al.(2019)Nguyen, Achille, Lam, Hassner, Mahadevan, and
  Soatto]{Nguyen2019}
Cuong~V. Nguyen, Alessandro Achille, Michael Lam, Tal Hassner, Vijay Mahadevan,
  and Stefano Soatto.
\newblock Toward understanding catastrophic forgetting in continual learning.
\newblock 2019, arXiv:1908.01091.

\bibitem[Lee et~al.(2021)Lee, Goldt, and Saxe]{Lee2021}
Sebastian Lee, Sebastian Goldt, and Andrew Saxe.
\newblock Continual learning in the teacher-student setup: Impact of task
  similarity.
\newblock In {\em {Proceedings of the 38th International Conference on Machine
  Learning, ICML}}, volume 139, pages 6109--6119, 2021.

\bibitem[Bengio et~al.(2009)Bengio, Louradour, Collobert, and
  Weston]{Bengio2009}
Yoshua Bengio, J\'{e}r\^{o}me Louradour, Ronan Collobert, and Jason Weston.
\newblock Curriculum learning.
\newblock In {\em Proceedings of the 26th Annual International Conference on
  Machine Learning, ICML}, page 41–48, 2009.

\bibitem[Hacohen and Weinshall(2019)]{Hacohen2019}
Guy Hacohen and Daphna Weinshall.
\newblock On the power of curriculum learning in training deep networks.
\newblock In {\em Proceedings of the 36th International Conference on Machine
  Learning, ICML}, volume~97, pages 2535--2544, 2019.

\bibitem[Weinshall et~al.(2018)Weinshall, Cohen, and Amir]{Weinshall2018}
Daphna Weinshall, Gad Cohen, and Dan Amir.
\newblock Curriculum learning by transfer learning: Theory and experiments with
  deep networks.
\newblock In {\em Proceedings of the 35th International Conference on Machine
  Learning, ICML}, volume~80, pages 5238--5246, 2018.

\bibitem[Mirzadeh et~al.(2020)Mirzadeh, Farajtabar, Pascanu, and
  Ghasemzadeh]{Mirzadeh2020}
Seyed~Iman Mirzadeh, Mehrdad Farajtabar, Razvan Pascanu, and Hassan
  Ghasemzadeh.
\newblock Understanding the role of training regimes in continual learning.
\newblock In {\em Advances in Neural Information Processing Systems},
  volume~34, 2020.

\bibitem[Sutton(1986)]{Sutton1986}
Richard~S. Sutton.
\newblock Two problems with backpropagation and other steepest-descent learning
  procedures for networks.
\newblock In {\em {Proceedings of the 8th Annual Conference of the Cognitive
  Science Society}}, pages 823--831, 1986.

\bibitem[McCloskey and Cohen(1989)]{McCloskey1989}
Michael McCloskey and Neal~J. Cohen.
\newblock Catastrophic interference in connectionist networks: The sequential
  learning problem.
\newblock {\em {Psychology of Learning and Motivation}}, 24:\penalty0 109--165,
  1989.

\bibitem[Ratcliff(1990)]{Ratcliff1990}
Roger Ratcliff.
\newblock Connectionist models of recognition memory: Constraints imposed by
  learning and forgetting functions.
\newblock {\em {Psychological Review}}, 97\penalty0 (2):\penalty0 285--308,
  1990.

\bibitem[French(1993)]{French1993}
Robert~M. French.
\newblock Using semi-distributed representations to overcome catastrophic
  forgetting in connectionist networks.
\newblock In {\em {Proceedings of the AAAI Spring Symposium}}, pages 70--77,
  1993.

\bibitem[Kirkpatrick et~al.(2017)Kirkpatrick, Pascanu, Rabinowitz, Veness,
  Desjardins, Rusu, Milan, Quan, Ramalho, Grabska-Barwinska, Hassabis, Clopath,
  Kumaran, and Hadsell]{Kirkpatrick2017}
James Kirkpatrick, Razvan Pascanu, Neil Rabinowitz, Joel Veness, Guillaume
  Desjardins, Andrei~A. Rusu, Kieran Milan, John Quan, Tiago Ramalho, Agnieszka
  Grabska-Barwinska, Demis Hassabis, Claudia Clopath, Dharshan Kumaran, and
  Raia Hadsell.
\newblock Overcoming catastrophic forgetting in neural networks.
\newblock {\em {Proceedings of the National Academy of Sciences of the United
  States of America}}, 114\penalty0 (13):\penalty0 3521--3526, 2017.

\bibitem[Zenke et~al.(2017)Zenke, Poole, and Ganguli]{Zenke2017}
Friedemann Zenke, Ben Poole, and Surya Ganguli.
\newblock Continual learning through synaptic intelligence.
\newblock In {\em {Proceedings of the 34th International Conference on Machine
  Learning, ICML}}, pages 3987--3995, 2017.

\bibitem[Kemker et~al.(2018)Kemker, McClure, Abitino, Hayes, and
  Kanan]{Kemker2018}
Ronald Kemker, Marc McClure, Angelina Abitino, Tyler~L. Hayes, and Christopher
  Kanan.
\newblock Measuring catastrophic forgetting in neural networks.
\newblock In {\em Proceedings of the 32nd AAAI Conference on Artificial
  Intelligence}, pages 3390--3398, 2018.

\bibitem[Goodrich and Arel(2014)]{Goodrich2014}
Ben Goodrich and Itamar Arel.
\newblock Unsupervised neuron selection for mitigating catastrophic forgetting
  in neural networks.
\newblock In {\em {Proceedings of the 57th IEEE International Midwest Symposium
  on Circuits and Systems}}, pages 997--1000, 2014.

\bibitem[Robins(1995)]{Robins1995}
Anthony Robins.
\newblock Catastrophic forgetting, rehearsal and pseudorehearsal.
\newblock {\em {Connection Science}}, 7\penalty0 (2):\penalty0 123--146, 1995.

\bibitem[Goodfellow et~al.(2014)Goodfellow, Mirza, Xiao, Courville, and
  Bengio]{Goodfellow2014}
Ian~J. Goodfellow, Mehdi Mirza, Da~Xiao, Aaron Courville, and Yoshua Bengio.
\newblock An empirical investigation of catastrophic forgeting in
  gradient-based neural networks.
\newblock In {\em International Conference on Learning Representations, ICLR},
  2014.

\bibitem[Hochreiter and Schmidhuber(1997)]{Hochreiter1997}
Sepp Hochreiter and J{\"{u}}rgen Schmidhuber.
\newblock Flat minima.
\newblock {\em Neural Computation}, 9\penalty0 (1), 1997.

\bibitem[Li et~al.(2017)Li, Xu, Taylor, Studer, and Goldstein]{Li2017}
Hao Li, Zheng Xu, Gavin Taylor, Christoph Studer, and Tom Goldstein.
\newblock Visualizing the loss landscape of neural nets.
\newblock In {\em Advances in Neural Information Processing Systems}, 2017.

\bibitem[Jastrz{\k{e}}bski et~al.(2017)Jastrz{\k{e}}bski, Kenton, Arpit,
  Ballas, Fischer, Bengio, and Storkey]{Jastrzkebski2017}
Stanis{\l}aw Jastrz{\k{e}}bski, Zachary Kenton, Devansh Arpit, Nicolas Ballas,
  Asja Fischer, Yoshua Bengio, and Amos Storkey.
\newblock Three factors influencing minima in {SGD}.
\newblock 2017, arXiv:1711.04623.

\bibitem[Mehta et~al.(2021)Mehta, Patil, Chandar, and Strubell]{Mehta2021}
Sanket~Vaibhav Mehta, Darshan Patil, Sarath Chandar, and Emma Strubell.
\newblock An empirical investigation of the role of pre-training in lifelong
  learning.
\newblock 2021, arXiv:2112.09153.

\bibitem[Ramasesh et~al.(2022)Ramasesh, Lewkowycz, and Dyer]{Ramasesh2022}
Vinay~Venkatesh Ramasesh, Aitor Lewkowycz, and Ethan Dyer.
\newblock Effect of scale on catastrophic forgetting in neural networks.
\newblock In {\em International Conference on Learning Representations, ICLR},
  2022.

\bibitem[Elman(1993)]{Elman1993}
Jeffrey~L. Elman.
\newblock Learning and development in neural networks: the importance of
  starting small.
\newblock {\em Cognition}, 48\penalty0 (1):\penalty0 71--99, 1993.

\bibitem[Krueger and Dayan(2009)]{Krueger2009}
Kai~A. Krueger and Peter Dayan.
\newblock Flexible shaping: How learning in small steps helps.
\newblock {\em Cognition}, 110\penalty0 (3):\penalty0 380--394, 2009.

\bibitem[Kumar et~al.(2010)Kumar, Packer, and Koller]{Kumar2010}
M.~Kumar, Benjamin Packer, and Daphne Koller.
\newblock Self-paced learning for latent variable models.
\newblock In {\em Advances in Neural Information Processing Systems},
  volume~23, 2010.

\bibitem[Wu et~al.(2021)Wu, Dyer, and Neyshabur]{Wu2021}
Xiaoxia Wu, Ethan Dyer, and Behnam Neyshabur.
\newblock When do curricula work?
\newblock In {\em International Conference on Learning Representations, ICLR},
  2021.

\bibitem[Schaul et~al.(2015)Schaul, Quan, Antonoglou, and Silver]{Schaul2015}
Tom Schaul, John Quan, Ioannis Antonoglou, and David Silver.
\newblock Prioritized experience replay.
\newblock 2015, arXiv:1511.05952.

\bibitem[Zhao and Zhang(2015)]{Zhao2015}
Peilin Zhao and Tong Zhang.
\newblock Stochastic optimization with importance sampling for regularized loss
  minimization.
\newblock In {\em Proceedings of the 32nd International Conference on Machine
  Learning, ICML}, volume~37, 2015.

\bibitem[Katharopoulos and Fleuret(2018)]{Katharopoulos2018}
Angelos Katharopoulos and Fran{\c{c}}ois Fleuret.
\newblock Not all samples are created equal: Deep learning with importance
  sampling.
\newblock In {\em {Proceedings of the 35th International Conference on Machine
  Learning, ICML}}, pages 12--18, 2018.

\bibitem[Baldock et~al.(2021)Baldock, Maennel, and Neyshabur]{Baldock2021}
Robert J.~N. Baldock, Hartmut Maennel, and Behnam Neyshabur.
\newblock Deep learning through the lens of example difficulty.
\newblock 2021, arXiv:2106.09647.

\bibitem[Cohn(1996)]{Cohn1996}
David~A. Cohn.
\newblock Neural network exploration using optimal experiment design.
\newblock {\em Neural Networks}, 9\penalty0 (6):\penalty0 1071--1083, 1996.

\bibitem[Settles(2009)]{Settles2009}
Burr Settles.
\newblock Active learning literature survey.
\newblock Technical report, University of Wisconsin-Madison Department of
  Computer Sciences, 2009.

\bibitem[Graves et~al.(2017)Graves, Bellemare, Menick, Munos, and
  Kavukcuoglu]{Graves2017}
Alex Graves, Marc~G. Bellemare, Jacob Menick, R{\'e}mi Munos, and Koray
  Kavukcuoglu.
\newblock Automated curriculum learning for neural networks.
\newblock In {\em Proceedings of the 34th International Conference on Machine
  Learning, ICML}, volume~70, pages 1311--1320, 2017.

\bibitem[Caruana(1997)]{Caruana1997}
Rich Caruana.
\newblock Multitask learning.
\newblock {\em Machine learning}, 28\penalty0 (1):\penalty0 41--75, 1997.

\bibitem[Pentina et~al.(2015)Pentina, Sharmanska, and Lampert]{Pentina2015}
Anastasia Pentina, Viktoriia Sharmanska, and Christoph~H. Lampert.
\newblock Curriculum learning of multiple tasks.
\newblock In {\em 2015 IEEE Conference on Computer Vision and Pattern
  Recognition, CVPR}, pages 5492--5500, 2015.

\bibitem[Guo et~al.(2018)Guo, Haque, Huang, Yeung, and Fei-Fei]{Guo2018}
Michelle Guo, Albert Haque, De-An Huang, Serena Yeung, and Li~Fei-Fei.
\newblock Dynamic task prioritization for multitask learning.
\newblock In {\em Proceedings of the European Conference on Computer Vision,
  ECCV}, 2018.

\bibitem[Lad et~al.(2009)Lad, Ghani, Yang, and Kisiel]{Lad2009}
Abhimanyu Lad, Rayid Ghani, Yiming Yang, and Bryan Kisiel.
\newblock Toward optimal ordering of prediction tasks.
\newblock In {\em Proceedings of the 2009 SIAM International Conference on Data
  Mining (SDM)}, pages 884--893, 2009.

\bibitem[Chaudhry et~al.(2018)Chaudhry, Dokania, Ajanthan, and
  Torr]{Chaudhry2018a}
Arslan Chaudhry, Puneet~K. Dokania, Thalaiyasingam Ajanthan, and Philip H.~S.
  Torr.
\newblock Riemannian walk for incremental learning: Understanding forgetting
  and intransigence.
\newblock In {\em Proceedings of the European Conference on Computer Vision,
  ECCV}, September 2018.

\bibitem[Chaudhry et~al.(2019)Chaudhry, Ranzato, Rohrbach, and
  Elhoseiny]{Chaudhry2018b}
Arslan Chaudhry, Marc’Aurelio Ranzato, Marcus Rohrbach, and Mohamed
  Elhoseiny.
\newblock Efficient lifelong learning with a-{GEM}.
\newblock In {\em International Conference on Learning Representations, ICLR},
  2019.

\bibitem[Chaudhry et~al.(2021)Chaudhry, Gordo, Dokania, Torr, and
  Lopez-Paz]{Chaudhry2021}
Arslan Chaudhry, Albert Gordo, Puneet Dokania, Philip Torr, and David
  Lopez-Paz.
\newblock Using hindsight to anchor past knowledge in continual learning.
\newblock volume~35, pages 6993--7001, 2021.

\bibitem[Goodfellow et~al.(2012)Goodfellow, Le, Saxe, Lee, and
  Ng]{Goodfellow2012}
Ian~J. Goodfellow, Quoc~V. Le, Andrew~M. Saxe, Honglak Lee, and Andrew~Y. Ng.
\newblock Measuring invariances in deep networks.
\newblock In {\em {Advances in Neural Information Processing Systems}},
  volume~22, pages 646--654, 2012.

\bibitem[Christianson(1992)]{Christianson1992}
Bruce Christianson.
\newblock Automatic hessians by reverse accumulation.
\newblock {\em IMA Journal of Numerical Analysis}, 12\penalty0 (2):\penalty0
  135--150, 04 1992.

\bibitem[LeCun et~al.(2010)LeCun, Cortes, and Burges]{LeCun2010}
Yann LeCun, Corinna Cortes, and Chris Burges.
\newblock {MNIST} handwritten digit database, 2010.

\bibitem[He et~al.(2016)He, Zhang, Ren, and Sun]{He2016}
Kaiming He, Xiangyu Zhang, Shaoqing Ren, and Jian Sun.
\newblock Deep residual learning for image recognition.
\newblock In {\em Proceedings of the IEEE Conference on Computer Vision and
  Pattern Recognition, CVPR}, 2016.

\bibitem[Krizhevsky(2009)]{Krizhevsky2009}
Alex Krizhevsky.
\newblock Learning multiple layers of features from tiny images.
\newblock Master's thesis, University of Toronto, 2009.

\bibitem[Paszke et~al.(2017)Paszke, Gross, Chintala, Chanan, Yang, DeVito, Lin,
  Desmaison, Antiga, and Lerer]{Paszke2017}
Adam Paszke, Sam Gross, Soumith Chintala, Gregory Chanan, Edward Yang, Zachary
  DeVito, Zeming Lin, Alban Desmaison, Luca Antiga, and Adam Lerer.
\newblock Automatic differentiation in {PyTorch}.
\newblock In {\em Autodiff Workshop, NeurIPS}, 2017.

\bibitem[Agarwal et~al.(2020)Agarwal, D'souza, and Hooker]{Agarwal2020}
Chirag Agarwal, Daniel D'souza, and Sara Hooker.
\newblock Estimating example difficulty using variance of gradients.
\newblock 2020, arXiv:2008.11600.

\bibitem[Kornblith et~al.(2019)Kornblith, Norouzi, Lee, and
  Hinton]{Kornblith2019}
Simon Kornblith, Mohammad Norouzi, Honglak Lee, and Geoffrey Hinton.
\newblock Similarity of neural network representations revisited.
\newblock 2019, arXiv:1905.00414.

\bibitem[Tsvetkov et~al.(2016)Tsvetkov, Faruqui, Ling, MacWhinney, and
  Dyer]{Tsvetkov2016}
Yulia Tsvetkov, Manaal Faruqui, Wang Ling, Brian MacWhinney, and Chris Dyer.
\newblock Learning the curriculum with {B}ayesian optimization for
  task-specific word representation learning.
\newblock In {\em Proceedings of the 54th Annual Meeting of the Association for
  Computational Linguistics}, volume~1, pages 130--139, 2016.

\bibitem[Javed and White(2019)]{Javed2019}
Khurram Javed and Martha White.
\newblock Meta-learning representations for continual learning.
\newblock In {\em Advances in Neural Information Processing Systems},
  volume~32, 2019.

\bibitem[Beaulieu et~al.(2020)Beaulieu, Frati, Miconi, Lehman, Stanley, Clune,
  and Cheney]{Beaulieu2020}
Shawn Beaulieu, Lapo Frati, Thomas Miconi, Joel Lehman, Kenneth~O. Stanley,
  Jeff Clune, and Nick Cheney.
\newblock Learning to continually learn.
\newblock In {\em Frontiers in Artificial Intelligence and Applications}, ECAI,
  pages 992--1001, 2020.

\end{thebibliography}
\normalsize

\end{document}